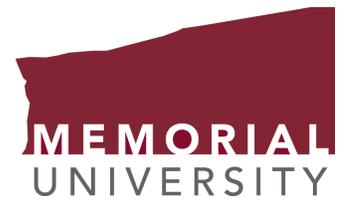

# AugmentTRAJ: A framework for point-based trajectory data augmentation

# Haranwala, Yaksh Jayeshkumar

Department of Computer Science

Memorial University of Newfoundland and Labrador

**Supervised by Dr. Amilcar Soares**

# Abstract

Data augmentation has emerged as a powerful technique in machine learning, strengthening model robustness while mitigating overfitting and underfitting issues by generating diverse synthetic data. Nevertheless, despite its success in other domains, data augmentation's potential remains largely untapped in mobility data analysis, primar- ily due to the intricate nature and unique format of trajectory data. Additionally, there is a lack of frameworks capable of point-wise data augmentation, which can reliably generate synthetic trajectories while preserving the inherent characteristics of the original data. To address these challenges, this research introduces AugmenTRAJ, an open-source Python3 framework designed explicitly for trajectory data augmentation. AugmenTRAJ offers a reliable and well-controlled approach to generating synthetic trajectories, thereby enabling the harnessing of data augmentation benefits in mobility analysis. This thesis presents a comprehensive overview of the methodologies employed in developing AugmenTRAJ and showcases the various data augmentation techniques available within the framework. AugmenTRAJ opens new possibilities for enhancing mobility data analysis models' performance and generalization capabilities by providing researchers with a practical and versatile tool for augmenting trajectory data. Its user-friendly implementation in Python3 facilitates easy integration into existing workflows, offering the community an accessible resource to leverage the full potential of data augmentation in trajectory-based applications.

# Acknowledgements

– Thank you to my supervisor, Dr. Amilcar Soares, for their advice and guidance through this research, for helping make this dissertation possible, and for being an amazing mentor, professor, and friend throughout my time at Memorial University.

– I would also like to thank Nicholas Jesperson for initiating the work for trajectory data augmentation and helping develop the initial prototype of AugmenTRAJ.

To my Family and Yesha, thank you for your unwavering support.

# Table of contents

# List of tables

# List of figures

# Chapter 1

# Introduction

## 1.1 Motivation and Significance

Over the past few years, research interest in movement and trajectory analysis has increased considerably because researchers are looking to aid fields such as transporta- tion planning [7] and its dynamics [35], livestock monitoring [9], and robotics [8], to name a few. Trajectory data captures the change in the position of an object relative to time. The most common sources of collection of such data are devices such as AIS sensors, GPS sensors, and mobile devices [18], which rely heavily on internet connectivity. Several tasks are necessary to properly work with trajectory data in a data mining setup, including: (i) data fusion [33, 21]; (ii) compression [10, 22];(iii) seg- mentation [34, 19, 6]; (iv) classification [12, 5]; (v) clustering [10, 42]; and (vi) outlier detection [11, 4, 2] to name a few. Furthermore, the collection of trajectory data has also raised some privacy concerns regarding tracking human subjects and their movements [40]. As a result, researchers often find it difficult to collect high-quality and sufficiently large datasets required for data-intensive tasks such as machine learning and data mining.

A popular way of dealing with the lack of data in many computer science domains is to generate synthetic data by performing data augmentation on existing real data. Data augmentation is the process of generating synthetic data by applying transfor- mations to the original data. Data augmentation has been proven highly effective in computer vision for object detection and image classification tasks [28, 30]. Not only have the data augmentation techniques been shown to increase the efficiency of the machine learning models, but they have also been shown to make the machine learning models more robust. Furthermore, data augmentation has been proven as an excellent alternative to enlarge datasets, cost-effectively using neural networks such as GANs (generative adaptive networks)[25].

To benefit from this idea, this work introduces AugmenTRAJ, a state-of-the-art and novel trajectory data augmentation package in python, that provides a collection of data augmentation techniques for trajectory data. AugmenTRAJ aims to provide the researchers with a streamlined process of augmenting trajectory data using popular Python 3 tools such as pandas [38] and PTRAIL [14, 15]. In summary, this thesis contributes with the following:

- AugmenTRAJ framework written in Python 3 that uses high Object Oriented Programming standards to allow for easy extension and ease of use in various environments.

- The implementation of AugmenTRAJ framework consists of the following packages:

    1. Candidate Trajectory Selection Strategies
        (a) Random Trajectory Selection
            – Randomly select a proportion of trajectories from the training dataset as augmentation candidates.
        (b) Proportional Trajectory Selection
            – Select an equal proportion of trajectories from each representative class in the training dataset as augmentation candidates.
        (c) Length-Based Trajectory Selection
            – Select a given proportion of trajectories that are shortest in length from the training dataset as augmentation candidates.
        (d) Representative Trajectory Selection

- Select the trajectories from the training dataset whose segment form statistics fall within a user defined range of the entire training dataset's statistics.

2. Trajectory Point Modification Strategies

   (a) In-circle Trajectory Point Modification Technique
   - Create the new point by considering a circular region around the current point and selecting the new point on the circumference of the circular region in a random direction.

   (b) On-circle Trajectory Point Modification Technique
   - Create the new point by considering a circular region around the current point and selecting the new point within the radius of the circular region in a random direction and at a random distance from the center.

   (c) Trajectory Point Stretching Modification Technique
   - Create the new point by shifting the current point by a user-given distance (in meters) and a user-given direction.

   (d) Point Dropping Modification Technique
   - Create the new trajectory by dropping the points from the original trajectory with a user-given probability.

- A subpackage of testing utilities that contains functions that can be used to set up the testing framework detailed in Section 3.4.

- The strategies provided by the framework have been tested extensively for cor- rectness and accuracy using *three* datasets containing trajectory data from var- ious domains, namely the Starkey [39] dataset for animal movement tracking, Geolife [43] and Traffic [14] dataset for transportation analysis.

# Chapter 2

# Literature Review

## 2.1 Overview of Data Augmentation Techniques

In recent years, data augmentation has gained much traction in the fields such as image processing, speech and audio processing, medical image processing, etc. As a result, immensely popular tools such as PyTorch [26] and TensorFlow [1] have embedded a dedicated sub-package for image data augmentation within their deep learning frameworks.

Generally, data augmentation techniques can be classified into two separate categories based on the time when augmentation is done in the machine learning pipeline that can be data augmentation during data preprocessing step 2.1.1 or data augmentation during model training step 2.1.2.

### 2.1.1 Data Augmentation During Data Preprocessing Step

Data augmentation during preprocessing is a very popular strategy due to its relative ease of use compared to active learning data augmentation. Furthermore, there are a wide variety of data augmentation techniques for various data domains, which further aids the popularity of data augmentation during preprocessing. According to [17], such techniques are:

1. Geometric Transformations

    - Geometric transformations for data augmentation include techniques such as data jittering, image cropping, image flipping, color distortion, geometric rotation, projection, and so on [23].

- Geometric transformations are generally used in image and video process- ing and are often very useful in the field of medical image processing to generate synthetic data due to the limited availability of data because of privacy reasons.

2. Fourier Transformations

   - Fourier transformations are another immensely popular data augmentation technique primarily used in the image and sound processing domain.
   - Fourier transforms are generally applied by converting the image from the spatial domain to Fourier domains and then by applying methods such as Fast Fourier Transform and Gaussian Noise injection to generate a con- trolled amount of variance in the data [25].

3. Time Series Augmentation

   - Time-series data is a collection of data points collected at certain time intervals and ordered chronologically along with the flow of time. Time- series data has become widely popular recently in domains such as financial analysis (stock markets and cryptocurrencies), livestock monitoring, and website usage monitoring.
   - In recent times, Time-series data has gained a lot of traction. Techniques such as time warping, slicing, permutation, and interpolation have become widely popular [25].

4. Many other data augmentation techniques specific to domains have become immensely popular due to recent advancements in machine learning and artificial intelligence [13] but are not that relevant for the purposes of this work.

## 2.1.2 Data Augmentation During Model Training

With the recent innovations and advancements in the field of computational capacity, along with the availability of faster computational chips, neural networks have become a popular alternative for generating synthetic data during the training phase of the model. Techniques such as Generative Adaptive Networks (GANs), Encoder-Decoder Networks have become very popular recently.

1. Generative Adversarial Networks (GANs)
    - Generative Adversarial Networks often comprise a combination of the Gen- erator Network and the Discriminator network. The main aim of the generator network is to create samples of synthetic data, whereas the discriminator network tries to tell apart the real and the synthetic data [37].
    - As Wang et al. [37] have mentioned, the effectiveness of Generative Adversarial Networks stems from their ability to estimate the distribution of the given data and generate synthetic data from it.
    - As a result, several Generative Adversarial Network architectures such as DCGAN [29] (Deep Convolutional Generative Adversarial Network) and StyleGAN [20] for computer vision applications andTimeGAN [41] for time- series applications have become immensely popular.

2. Encoder-Decoder Networks
    - Encoder-Decoder networks are another popular alternative with an architecture similar to the Generative Adversarial Networks.
    - As the name suggests, encoder-decoder networks consist of two distinct structures; the encoder network, which transforms reduces the higher dimensional data into a lower dimension, whereas the decoder network tries to reconstruct the lower dimensional data into the original higher dimensional [16].

- As a result of the above, encoder-decoder networks seldom generate purely synthetic data; rather, the data generated by encoder-decoder networks is usually a combination of features learned during the decoder process.

3. Apart from the Generative Adversarial Networks and Encoder-Decoder networks, other methods such as Rule-Based data generation techniques such as Procedural Content Generation (PCG) [32] are also very popular in the gaming context. Apart from that, Bae et al. (2018) [3] have also used Perlin Noise to perform data augmentation on HRCT (High-Resolution Computed Tomogra- phy) images in the context of medical image analysis.

### 2.1.3 Conclusion

Trajectory data is usually stored as points spaced apart in time and location data at each point. Due to such intricate nature of the data, geometric augmentation methods cannot be applied to trajectory data. Furthermore, Fourier transformations are seldom used outside of the image and wave analysis domain and are unsuited for the trajectory analysis domain. However, time-series data closely resemble trajectory data because trajectory data is similar to time-series data as it has points spread apart in time. As a result, augmentation methods in the time-series domain can be modified to be applied to the trajectory domain. Therefore in this work, we took inspiration from such techniques and modified them to suit the needs of the trajectory data analysis domain.

In the next section, we will discuss the novel data augmentation techniques we propose using AugmenTRAJ for trajectory data. Section 4 will discuss the effect of augmenting trajectory data using AugmenTRAJ on machine learning tasks such as classification. Finally, in Section 5, we will summarize the AugmenTRAJ package along with a brief discussion about the future steps for AugmenTRAJ.

# Chapter 3

8# Materials and Methods

Trajectory data is often represented in two formats, namely *point-based* format and *segmented-based* format [14]. In the point-based format, trajectory data is recorded at regular intervals with the location, i.e., latitude and longitude of the subject being a requirement. Other relevant data, such as altitude, time of the day, and data about the environment that the subject is in, are often stored with each point recorded. On the other hand, in the segment-based form of the trajectory data, the point-based form of the trajectory is usually divided into segments and represented as a row in the data table containing statistical values of the trajectory such as mean speed, total distance traveled, displacement from the start and so on. In a segment-based format, data is represented as one trajectory per row containing all the statistical data about the entire trajectory. Figure 3.1a and Figures 3.1b respectively show the point-based and the segment-based format of the trajectory data from the Starkey [39] animals dataset.

Data augmentation in trajectory and movement analysis has not gained popularity due to the complex nature of trajectory data. However, due to the sequential nature of trajectory data, several data augmentation techniques, such as noise introduction, shifting, and scaling that are popular for generic time-series data, can also be applied to trajectory data with necessary modifications. Furthermore, as described before, a segment-based trajectory only contains the statistical values of the trajectory as the data. As a result, a point-based format is preferred to perform augmentation on trajectory data. Another point to be highlighted here is that applying modifications or noise in the segment-based format could lead to invalid instances. For example

| traj_id | DateTime | lat | lon | StarkeyTime | GMDate | GMTime | LocDate | LocTime | RadNum | Species | UTME | UTMN | Year |
|---|---|---|---|---|---|---|---|---|---|---|---|---|---|
| 880109D01 | 1995-04-13 13:40:06 | 45.239682 | -118.533204 | 229902006 | 21:40:06 | 19950413 | 19950413 | 13:40:06 | 409 | 0 | 379662 | 5010734 | 95 |
| | 1995-04-15 12:16:15 | 45.250521 | -118.530438 | 230069775 | 20:16:15 | 19950415 | 19950415 | 12:16:15 | 409 | 0 | 379895 | 5011927 | 95 |
| | 1995-04-15 21:39:38 | 45.247943 | -118.541455 | 230103578 | 05:39:38 | 19950416 | 19950415 | 21:39:38 | 409 | 0 | 379039 | 5011656 | 95 |
| | 1995-04-16 03:32:14 | 45.247429 | -118.539530 | 230124734 | 11:32:14 | 19950416 | 19950416 | 03:32:14 | 409 | 0 | 379188 | 5011581 | 95 |
| | 1995-04-16 04:08:28 | 45.247117 | -118.542579 | 230126908 | 12:08:28 | 19950416 | 19950416 | 04:08:28 | 409 | 0 | 378938 | 5011567 | 95 |

(a) Point Based Trajectory Format

| traj_id | 10%_Distance | 25%_Distance | 50%_Distance | 75%_Distance | 90%_Distance | min_Distance | max_Distance | mean_Distance | std_Distance |
|---|---|---|---|---|---|---|---|---|---|
| 910313E37 | 30.022359 | 66.956787 | 149.611989 | 300.726867 | 632.269559 | 0.0 | 6034.207873 | 268.728974 | 398.901980 |
| 890424E08 | 42.384642 | 84.780370 | 177.223376 | 379.614004 | 759.311144 | 0.0 | 6043.490157 | 323.956468 | 446.249248 |
| 921228E06 | 42.390114 | 90.066813 | 200.849995 | 429.158672 | 890.194719 | 0.0 | 4623.347553 | 377.677630 | 517.650637 |
| 930304E16 | 42.384012 | 67.090258 | 153.068542 | 341.254645 | 690.640774 | 0.0 | 5864.020183 | 291.351804 | 396.220730 |
| 940110D01 | 30.022192 | 66.956668 | 149.612107 | 284.710015 | 510.276991 | 0.0 | 1975.611748 | 221.420080 | 235.093739 |

(b) Segment Based Trajectory Format

Figure 3.1: Trajectory Formats

when a moving object increases its speed, its acceleration should also be increased. If a data augmentation technique applies noise on these columns individually, we could create a situation where the speed increases while the acceleration decreases. In such a situation, we would be creating invalid data.

When augmenting data, the primary purpose is to generate samples close to the original data's distribution and represent the actual data very closely. Therefore, it is of utmost importance that we do not generate samples that can be easily discerned from the original data to work towards the end goal of improving the efficiency of the machine learning model. Trajectory data augmentation is affected by two processes, namely *selection* of trajectories to be augmented and *modification* techniques used to alter points in the original trajectory to generate the synthetic data.

To tackle the aforementioned challenges, we have developed several techniques to select trajectories to be augmented based on different criteria. Once the trajectories are selected, AugmenTRAJ allows users to augment them using several methods pro- vided out of the box. In the following subsection, we will describe all the selection and point-modification techniques provided by AugmenTRAJ.

# 3.1 Augmentation Candidate Trajectory Selection Techniques in AugmenTRAJ

The selection of trajectories to be augmented plays a significant role in the trajectory data augmentation procedure as it can greatly affect the efficiency of the machine learning model. Furthermore, when the candidates are chosen correctly, it often helps balance the skewed data toward one of the instances and classes and can help the model discern the classes better. On the other hand, when the dataset is balanced, but the trajectories within each class are very similar, the machine learning model is prone to overfitting the training data and performing much worse on the testing data. In such cases, selecting and augmenting the right trajectories helps introduce variation in the samples to help the model learn better and not overshoot the local minima.

In machine learning, the model is usually trained on a subset of the original data, which is called the *training data*, whereas the efficiency inference of the model is done using a subset of data called the *testing data* that the model has not seen in the training stages. In the following subsections, we will describe all the trajectory selection methods that AugmenTRAJ provides out-of-the-box, but it is to be kept in mind that the selection of trajectories as well as the modification is made to the training data exclusively to increase the size, quality, and variation of the data to help training a better model. Hence, in the following sections, when we mention selection and modification of data, it is done exclusively on the training data. In AugmenTRAJ, we have developed the following four techniques for selecting the candidates for augmentation:

## 3.1.1 Random Trajectory Selection

As the naming suggests, in the random trajectory selection method, the trajectories to be augmented are selected randomly from the training dataset. The users can control the proportion of the training data they want to augment. For instance, if the training dataset has 100 trajectories, and the user wants to augment 20% of the data, 20 trajectories will be selected randomly from the dataset. Furthermore, the users can control the randomness for the reproducibility of results by passing in a seed to the random selection.

Random trajectory selection is often useful as the stepping-stone to determine if a different trajectory selection is warranted. Since random trajectory selection does not look at the distribution of the data, it is possible that it may select trajectories from a class that is a very large proportion of the data and may end up increasing the skewness of the data. In such cases, Dataset balancing can greatly help the model improve.

### 3.1.2   Proportional Trajectory Selection

In a proportional selection of trajectories, the user is allowed to select what proportion of trajectories are selected from each class of the data points for augmentation. For example, consider a dataset that has the following structure: (i) Class A: 20 trajec- tories; (ii) Class B: 5 trajectories. If the user wants 20% of trajectories from each class, then **3 trajectories** are selected from class A, and **1 trajectory** is selected from class B. Figure 3.2 depicts the visual representation of this process.

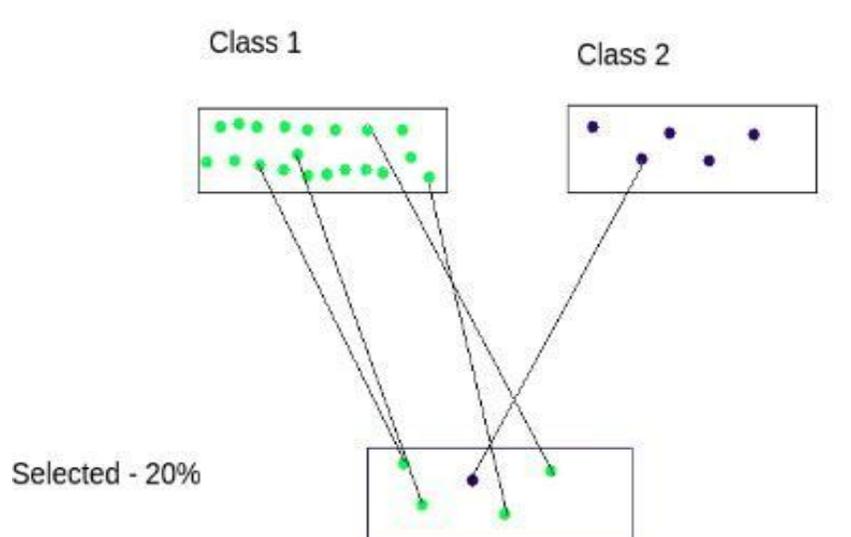

Figure 3.2: Proportional Selection Technique in AugmenTRAJ

Proportional trajectory selection is an excellent choice when we know that the dataset is well-balanced and each class is represented with plenty of samples. In such datasets, augmentation of proportionally selected trajectories can increase the number of samples for each class and introduce further variations in the data for each class to better assist the model in training.

## 3.1.3 Length-Based Trajectory Selection

In the length-based trajectory selection method, a given proportion of the trajectories with the shortest length from the data are selected for augmentation. For instance, if a dataset has 100 trajectories and the user wants to select 20% of trajectories to be augmented, then the trajectories will be sorted according to their lengths in ascending order, and the 20 shortest trajectories will be selected for augmentation. Figure 3.3 depicts the fewest selection technique in AugmenTRAJ.

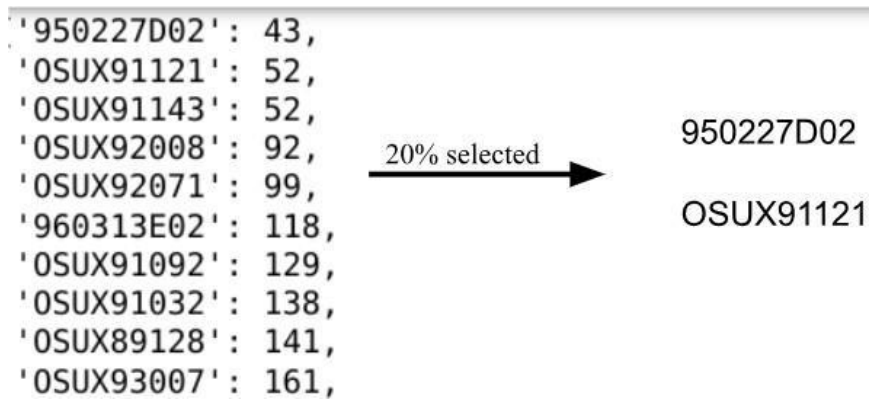

Figure 3.3: Fewest Selection Technique in AugmenTRAJ

As we have mentioned before, trajectory data is often difficult to collect, maintain, and store, so we often get extremely short trajectories. Machine learning models usu- ally require a lot of good-quality data, and having shorter or incomplete trajectories can often lead to the model underfitting the data and performing worse on the test- ing set. In such cases, selecting the shortest trajectories and augmenting them can increase the size of the data and make the model more robust to variations in such smaller trajectories.

## 3.1.4 Representative Trajectory Selection

In the representative trajectory selection technique, the trajectories are selected based on the closeness of an individual trajectory's statistics to that of the entire dataset.

14To do so, first, the entire data is converted to segment-based form wherein each data row is one trajectory, and its statistics, such as mean, median, maximum, minimum, and so on, are calculated for kinematic features such as distance, displacement, speed, acceleration, and jerk. This is easily achieved using a single command available in the PTRAIL [14] package for trajectory data processing. Once the statistics for the entire dataset are calculated, each trajectory's statistics are compared with that of the entire dataset, and if the user-given proportion of trajectory statistics falls within the user-given tolerance level, then it is selected for augmentation. It must be noted how each selection mechanism provides the user with fine-grained control of the process, thereby following very high software engineering principles.

The representative selection technique works great in getting data distribution closer to a bell-curve representation. It is generally applicable when the dataset is well-balanced and representative classes well represented by increasing the number of samples in the training data. However, depending on the user-given tolerance for selection, representative selection may select all the trajectories in the dataset if the dataset is fairly balanced. Hence, representative selection should be tried with a few values of tolerance and select the most appropriate one where it selects a good number of trajectories but does not select all or most of them as it could slow down the augmentation process significantly and have little to no effect on the efficiency of the model.

## 3.2 Point Modification Techniques for Generating Synthetic Trajectories in AugmenTRAJ

Once the desired set of trajectories has been selected to be augmented, the next step is to modify the points within each selected trajectory to generate the synthetic trajectories. When augmenting data, it is useful to restrict the user-given parameters to the augmentation methods at sensible levels, and as such, it is always good to try out several techniques for data before choosing the best one because there is no one-size-fits-all solution when it comes to data augmentation. As a result, in AugmenTRAJ, we have developed the following four techniques for point modification:

## 3.2.1 On-Circle Modification

In the on-circle point modification technique, a point is modified using the following technique:

- Select a circle of a given radius around the point in the original trajectory. The radius is calculated using 10% of the distance between the current point and the next point in the trajectory.

- Randomly select a degree for modifications and place the new point on the circle's circumference in the direction of the selected degree.

- Figure 3.4 demonstrates the on-circle point modification technique in Augmen- TRAJ.

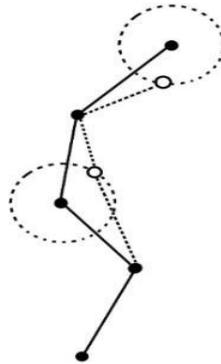

Figure 3.4: On-Circle Point Modification Technique in AugmenTRAJ

In this modification technique, each point's location, i.e., latitude and longitude, of the selected trajectory is modified using the process above. However, the trajectory will have the same time intervals as the original trajectory as we only modify the location of the points, and the other features in the data are kept unmodified.

A significant advantage of this technique is that the variance induced in the data is not huge, as the synthetic trajectories generated represent the original trajectories to some degree due to the nature of the modification technique. As a result, this technique is very useful when it is known that the data already has enough variations in each represented class. This technique will help increase the data samples with a limited variance.

### 3.2.2 In-Circle Modification

The in-circle point modification technique is similar to the on-circle point modification technique, and the point is modified using the following technique:

- Randomly select a distance in a circular region around the current point where the radius of that region is smaller than the distance between the current point and the next.

- Randomly select a degree and place the point at the aforementioned distance in the direction of the selected degree.

- Figure 3.5 demonstrates the in-circle point modification technique in Augmen- TRAJ.

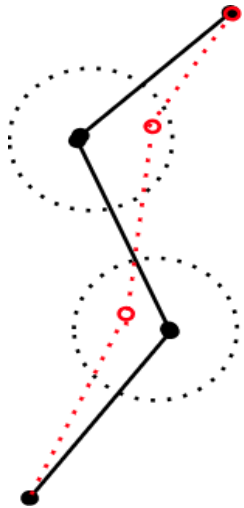

Figure 3.5: In-Circle Point Modification Technique in AugmenTRAJ

Similar to the on-circle point modification technique, only the points' locations, i.e., latitude and longitude, are modified for the entire trajectory with timestamps and other features remaining intact. However, as compared to the on-circle point modification technique, where only a limited degree of modifications to the point can be made due to the restriction of the new point being on the circumference of the selected circular region, the in-circle point modification technique allows for a higher degree of flexibility in terms of distance of the point as well as the direction as compared to the original point.

The primary advantage of the in-circle method is that it allows for a higher level of variance to be introduced to the data compared to the on-circle modification method. As a result, this method is very useful when we want to introduce some amount of variance in the dataset along with increasing the number of samples of the data.

### 3.2.3 Point Stretching Modification

The basic idea behind stretching modification is moving each point in the user-specified direction by a user-specified magnitude to generate synthetic trajectories. The points are modified using the following technique:

- Based on the maximum allowed user-given distance, calculate the maximum latitudes and longitudes in each direction on a straight line that passes through the current point illustrated in Figure 3.6.

- Once the bounds are calculated, the new point is calculated based on the user- given method. The user can choose one of the following methods for selecting the new point:

    1. Minimum Distance Point
        – Always select the point that is on the *minimum* side as displayed in Figure 3.6.
    2. Maximum Distance Point
        – Always select the point that is on the *maximum* side as displayed in Figure 3.6.
    3. Minimum/Maximum Distance Point Randomly
        – Randomly select either the maximum or the minimum point in the new trajectory as displayed in Figure 3.6.
    4. Random Point between Minimum and Maximum
        – Randomly select a point between the minimum and the maximum bounds in Figure 3.6.

The point stretching method gives the user fine-grained control to the user in terms of choosing how the synthetic trajectory will be generated. Not only does it allow the user to select where the new point will be in terms of the distance, but it also allows the user to control the direction in which the new point will be generated as compared to the in-circle and on-circle methods where the user has no control over the direction and the distance of the new point.

The stretching modification method is generally useful in most situations as the user can control the variance induced in the data using the distance and the point selection method. As a result, the user can either generate more samples with minimal variance or more variance to enrich the data.

### 3.2.4 Point Dropping Modification

In the point-dropping method, synthetic trajectories are generated by dropping points based on a user-given probability. The process is illustrated in Figure 3.7.

The point dropping drops the point from the trajectory, due to which the resultant trajectory is prone to abrupt jumps in the path of the object. As a result, the point- dropping method can potentially introduce the most variance in the data samples. Therefore, it is upon the user to control the probability of dropping the points, hence controlling the variance in the new dataset. This method is especially useful when the data samples in the same class and among different classes are very similar as it can introduce the required variance to help the modern learn the distinction among the classes better and not underfit in such cases.

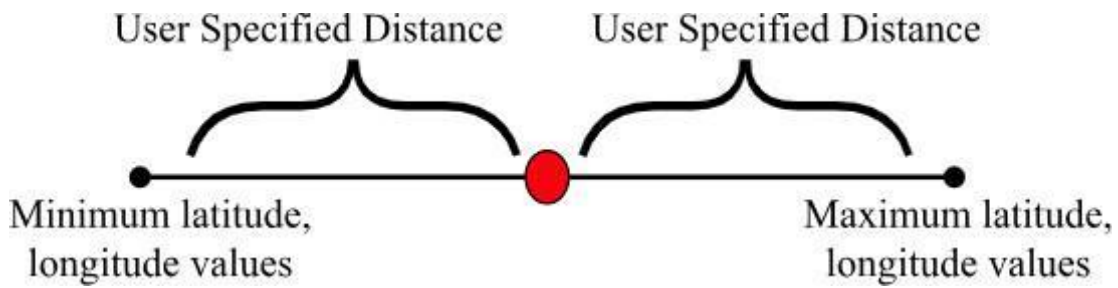

Figure 3.6: Minimum and Maximum Latitude-Longitude Calculation

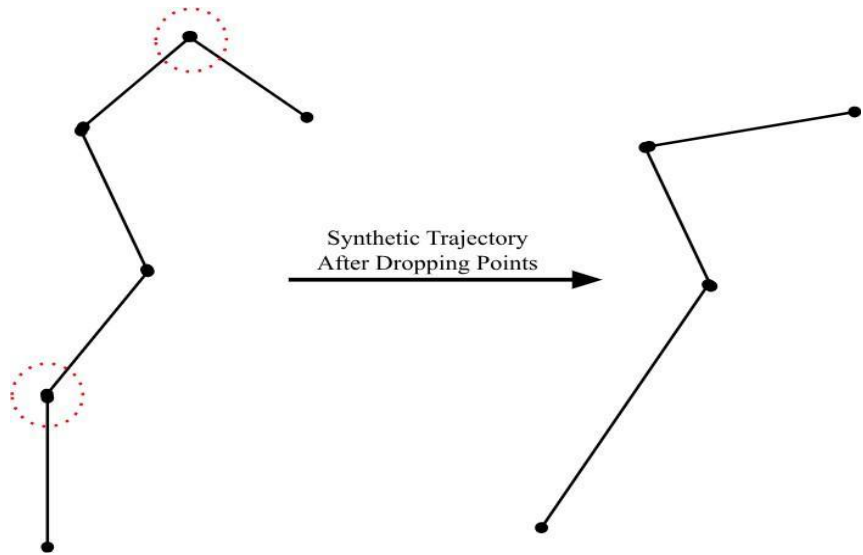

Figure 3.7: Point Dropping Modification Technique in AugmenTRAJ

## 3.3 Dataset Balancing Techniques in AugmenTRAJ

As we have briefly mentioned before, collection and storage of trajectory data is often a difficult and tedious task plagued by many issues such as loss of connection, faulty sensors, etc. Furthermore, the collection of movement data for humans also brings privacy concerns. With the world being connected more and more with each other through the internet, having a human's frequently traveled path along with its metadata can be highly risky for that person. Due to these reasons, movement datasets are often restricted to tracking animals, shifts, aircraft, etc.

Apart from that, when data for the movement of humans is available, the mode of transport is often modes of public transport such as trains, subways, buses, etc., and not personal transportation vehicles such as cars and bikes. In such cases, the machine learning model often underfits the data and performs worse at accurately classifying the trajectories since it does not have enough data for the classes with fewer data points. Furthermore, it is generally seen that machine learning models perform very well when the dataset is balanced or near-balanced, with each representative class having an equal number of samples. As a result, to deal with the aforementioned challenge and to take advantage of a balanced dataset, AugmenTRAJ provides techniques for the balanced dataset with a single method call. The functioning of dataset balancing techniques is as follows

Consider a dataset with the following balance of classes: (i) Class 1: 50 trajectories; (ii) Class 2: 100 trajectories; (iii) Class 3: 75 trajectories, totaling 225 trajectories. If the user wants to balance the dataset, then an easy approach would be to augment samples in classes 1 and 3 such that each class finally has 100 samples. However, that will mean that classes 1 and 3 will have several synthetic trajectories and are induced into their distributions. However, it is generally ideal to induce variance into all the representative classes in the dataset using augmentation.

Therefore, in AugmenTRAJ, the user is asked to specify a target number of trajectories for each class as a multiplier of the class with the highest number of samples. To simplify this, continuing with our example dataset above, if the user specifies a multiplier of *1.1*, then we would have a total of 110 samples for each class because our Class 2 has the maximum number of samples with 100 samples and we will get *100 * 1.1 = 110* based on the maximum target multiplier of 1.1. Hence, our final dataset will look as follows: (i) Class 1: 110 trajectories; (ii) Class 2: 110 trajecto- ries; (iii) Class 3: 110 trajectories; a total of 330 trajectories. Since AugmenTRAJ functionalities are very flexible, all the techniques for modifying points are available for augmentation while balancing the datasets.

## 3.4  Experimental Setup

To test the level of impact of synthetic data generation using AugmenTRAJ, we have implemented a setup wherein we used various machine learning models such as Random Forest, Gradient Boosting Classifier, Support Vector Classifier, and Decision Tree classifier to predict what class the object belongs to using the trajectory data as the input. The framework that we setup for testing is as follows:

1. Select 80% of the trajectories from the given dataset that will be used as the training dataset, and the rest 20% of the trajectories will be set aside for the testing dataset.

2. Once the training data is selected, pre-select and store the augmentation candi- dates in a dictionary using each of the methods described in Table A.1.

3. Next, using each of the dataset balancing techniques, balance the datasets and store them in the aforementioned dictionary.

4. Once the dictionary above is created with the balanced datasets and augmenta- tion candidates using various techniques, we used 20 seeds (we used digits of Pi after the decimal point) to train various models for the training dataset without augmentation and then augmenting datasets using each of the techniques de- scribed in Table A.2. It should be noted that while performing augmentation, we augmented each selected candidate trajectory *three* times to generate three distinct samples based on the original trajectory.

5. For each model, we calculate the f1 score and accuracy and compare those metrics with the dataset without augmentation and calculate the increase in accuracy and f1 score induced using augmentation.

6. Figure 3.8 illustrates the testing framework described above.

Using this pipeline[1], we trained 63 machine learning models (i.e., combining dis- tinct ML models, trajectory selection, and trajectory point modification strategies) for different variants of training data and compared them with the results of the models trained with un-augmented training data. Following core software engineering principles, the framework for testing the augmentation strategies have been engineered to be used with new datasets and different machine learning models available in the scikit-learn [27] python package. In the following section, we will discuss the results that we obtained by using the above framework on the Starkey animals dataset [39], a subset of Microsoft Geolife database [43] and the Traffic dataset available as one of the datasets in the PTRAIL library [14].

Finally, it is important to describe the metrics we used to compare the accuracy of models in our experiments. To determine the enhancements caused by augmentation techniques, the following metrics were used:

1. F1 Score

    - Mathematically, the f1 score is defined as the mean of the model's precision and recall.

$$F1 - score = \frac{2 * precision * recall}{precision + recall} \qquad (3.1)$$

---
[1]Example Jupyter notebook containing the testing described above: Link.

```python
from ptrail.core.Datasets import Datasets
from ptrail.features.kinematic_features import KinematicFeatures
from sklearn.ensemble import GradientBoostingClassifier
from sklearn.metrics import f1_score, accuracy_score
from sklearn.svm import SVC
from sklearn.tree import DecisionTreeClassifier
from src.utils.general_utils import Utilities
from TestUtils.test_utils import TestUtils
from TestUtils.Keys import *

# Get the 20 seed values that we are going to use.
seed_generator = Utilities.generate_pi_seed(20)
seed_vals = [next(seed_generator) for i in range(20)]
final_results = ["seed, strategy, model, accuracy, f1_score"]

# All our selection strategies.
select_strategies = [
    BASE, RANDOM_SELECTED, PROPORTIONAL_SELECTED, FEWEST_SELECTED, REPRESENTATIVE_SELECTED,
    BALANCED_ON, BALANCED_IN
]
augment_strategies = [ON, IN, DROP, STRETCH]
models = [GradientBoostingClassifier(), DecisionTreeClassifier(), SVC()]

for seed in seed_vals:
    # Get the iterable map for the seed.
    iter_map = TestUtils.get_iterable_map(ready_dataset, seed, 'vehicle_type')
    for select_strategy in select_strategies:
        for model in models:
            if select_strategy != BASE and 'balanced' not in select_strategy:
                for augment_strategy in augment_strategies:
                    train_x, train_y = TestUtils.select_correct_test_train_split(iter_map, select_strategy,
                                                                                  augment_strategy,
                                                                                  'vehicle_type', 3)
                    if (train_x is not None) and (train_y is not None):
                        # Fit the model and predict.
                        model.random_state = seed
                        model.fit(X=train_x, y=train_y)
                        pred_vals = model.predict(X=iter_map[TEST_X])

                        # Calculate the accuracy and f1 score.
                        acc = accuracy_score(y_true=iter_map[TEST_Y], y_pred=pred_vals)
                        score = f1_score(y_true=iter_map[TEST_Y], y_pred=pred_vals, average='weighted')
                        print(f"{seed}, {select_strategy} {augment_strategy}, {model.__class__.__name__}, {acc}, {score}")
                        final_results.append(f"{seed}, {select_strategy} {augment_strategy}, {model.__class__.__name__}, {acc}, {score}")
            else:
                train_x, train_y = TestUtils.select_correct_test_train_split(iter_map, select_strategy,
                                                                              BASE, 'vehicle_type', 3)
                if (train_x is not None) and (train_y is not None):
                    # Fit the model and predict.
                    model.random_state = seed
                    model.fit(X=train_x, y=train_y)
                    pred_vals = model.predict(X=iter_map[TEST_X])

                    # Calculate the accuracy and f1 score.
                    acc = accuracy_score(y_true=iter_map[TEST_Y], y_pred=pred_vals)
                    score = f1_score(y_true=iter_map[TEST_Y], y_pred=pred_vals, average='weighted')
                    if 'balanced' not in select_strategy:
                        print(f"{seed}, base, {model.__class__.__name__}, {acc}, {score}")
                        final_results.append(f"{seed}, base,'f' {model.__class__.__name__}, {acc}, {score}")
                    else:
                        print(f"{seed}, {select_strategy}, {model.__class__.__name__}, {acc}, {score}")
                        final_results.append(f"{seed}, {select_strategy},'f' {model.__class__.__name__}, {acc}, {score}")
```

Figure 3.8: Code Snippet Illustrating the Testing Framework For AugmenTRAJ

- The F1-score aims to combine two relatively simple metrics, precision and recall, to give a more complete score of how good the model is doing.

2. Accuracy

- Accuracy measures how many correct predictions the model makes out of all the predictions made.

$$Accuracy = \frac{Number\ of\ Correct\ Predictions}{Total\ Predictions} \qquad (3.2)$$

# Chapter 4

# Results and Discussion

This section presents the experiments' outcomes as outlined in Section 3.4. After training all models with different augmentation techniques, the performance improve- ment in accuracy and F1-score was assessed. Box plots were generated to facilitate comparison, showcasing the results for each machine learning model and augmenta- tion strategy combination using AugmenTRAJ. These box plots can be observed in Figures 4.1, 4.2, and 4.3.

As observed in Figures 4.1, 4.2, and 4.3, the implementation of augmentation strategies generally leads to a reduction in the error rate for all utilized datasets, signifying the potential of data enhancement in improving model performance. How- ever, it is crucial to acknowledge that augmentation does not invariably result in error rate improvement; in some cases, it may introduce higher data variance, consequently leading to underfitting.

Upon initial inspection, the box plots do not demonstrate a substantial increase in accuracy for the models trained in the classification tasks. Nonetheless, it's essential to recognize that effective machine-learning tasks require systematically exploring var- ious model configurations. In this study, we trained different machine learning models under different controlled conditions using 20 seed values and identified the optimal configurations for each model. The Table B.1 demonstrates maximum performance improvement attained with AugmenTRAJ's augmentation strategies for the geolife dataset

Figure 4.1: Geolife Dataset Box Plot

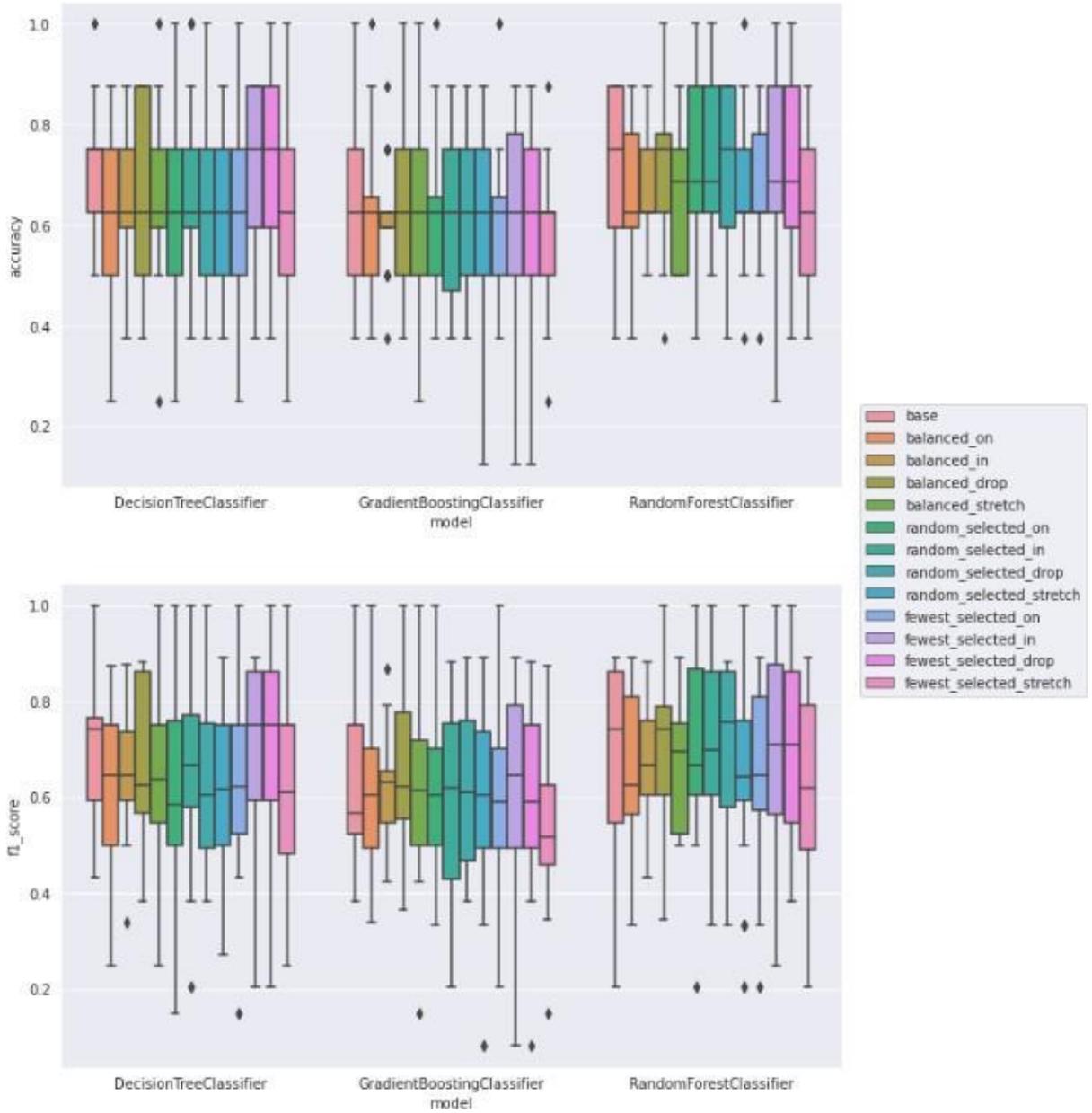

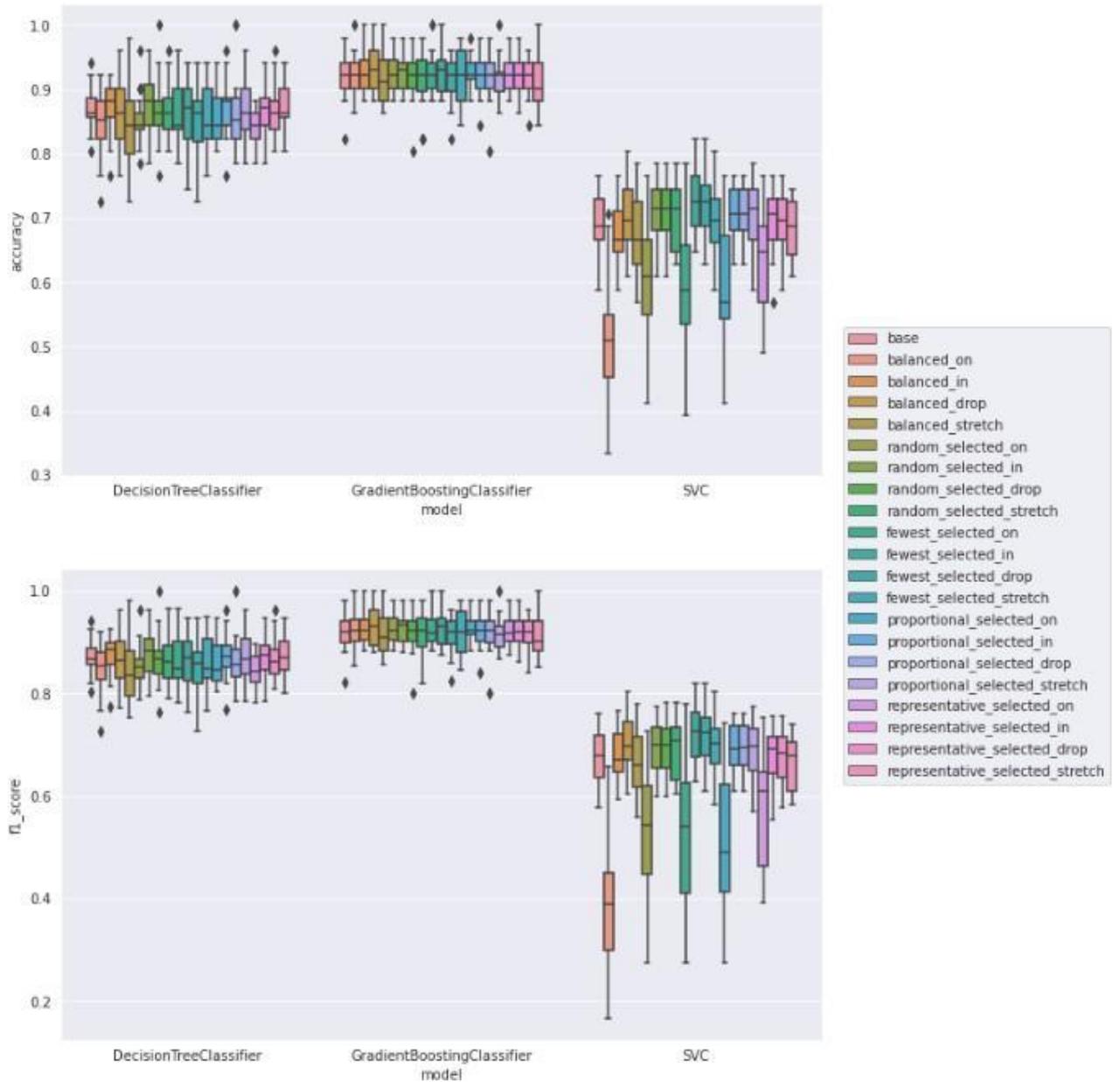

Figure 4.2: Starkey Dataset Box Plot

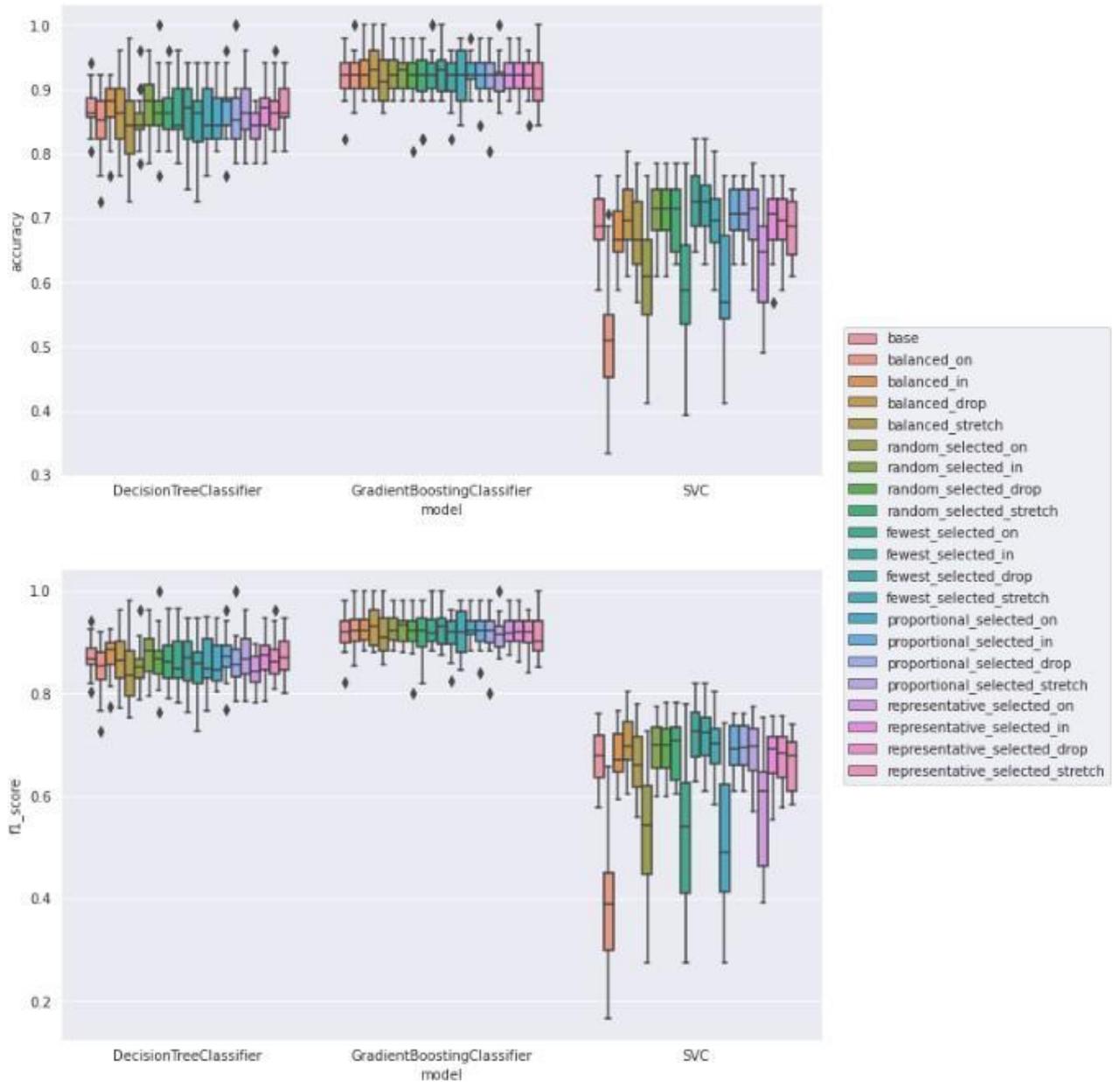

Figure 4.3: Traffic Dataset Box Plot

The controlled experiment with seed 7950 reveals striking improvements when us- ing augmentation strategies. The accuracy and f1-score for the base strategy, i.e., training data without augmentations, were extremely poor at 0.38 and 0.2, respectively. However, through the generation of synthetic data and dataset balancing using the point dropping method (Section 3.2.4), the accuracy and f1-score remarkably in- creased to 0.75 for both metrics. This emphasizes the significance of dataset balancing techniques, particularly in unbalanced class distributions within the geolife dataset. When the training data is skewed towards heavily represented classes, models tend to over-fit and under-perform on the testing dataset. Augmentation techniques address this issue by providing ample samples for each class, enhancing the training dataset's balance, and improving model performance. Additionally, the geolife dataset contains trajectories with very short lengths, posing challenges for machine learning models to identify meaningful patterns and classify them accurately. Augmenting shorter trajectories can significantly enhance the training dataset by providing more samples for the models to learn from. This, in turn, leads to substantial improvements in model performance on the testing set. Seed values 2643 and 4944 notably demonstrate such enhancements, where metrics improved by over 30%. AugmenTRAJ's augmentation strategies offer valuable solutions for addressing the intricacies of the geolife dataset, mitigating class imbalances, and handling trajectories of varying lengths. By doing so, the framework enables the models to learn more effectively and achieve significantly improved classification performance.

Shorten and Taghi M. [31], and Taylor and Nitschke [36] have shown a reduction of error rate in their models by about 1% whereas Moreno et al. [24] have shown an increase in their models' test accuracy of up to 5% by using data augmentation in the domain of image classification. However, the improvements are significantly lower than 51% improvement that has been showcased using AugmenTRAJ in trajec- tory classification under controlled conditions. This emphasizes how the techniques of data augmentation can greatly aid the machine learning tasks in the movement data analysis domain, and when used correctly, it can greatly help in training models that perform very well on such data. Tables B.2 and B.3 summarize the maximum im- provement resulting from augmentation techniques in AugmenTRAJ for the Starkey and the traffic dataset similar to the Table B.1

# Chapter 5

28# Conclusion

This study has introduced AugmenTRAJ, a framework for data augmentation in the movement data analysis domain. While data augmentation has been extensively explored in various domains, such as image and time series, its potential in movement analysis remains largely untapped. The challenges associated with collecting move- ment data often complicate data mining, necessitating a specialized solution to address these issues.This research aimed to develop a framework that effectively addresses the unique complexities of movement data analysis, allowing machine learning models to perform at par with their counterparts in other computer science domains.

Though data augmentation is not a panacea for all movement data analysis challenges, AugmenTRAJ's exceptional results have surpassed industry standards. This signifies the bright future of data augmentation in the movement data domain. Aug- menTRAJ's success opens up possibilities for advanced topics such as Generative Adversarial Networks (GANs) in movement data analysis, promising further advancements in the field. Furthermore, the availability of AugmenTRAJ as an open-source resource to developers worldwide is expected to drive increased interest and traction in data augmentation for movement data analysis. Collaboration and contributions from the community will enrich the framework and foster innovation in the field. While AugmenTRAJ represents a significant step forward, there is still much ground to cover in the movement data analysis domain. Future research should focus on enhancing the framework, exploring novel augmentation techniques, and fostering interdisciplinary collaborations to unlock the full potential of data augmentation in movement analysis.

In conclusion, AugmenTRAJ marks a pivotal contribution to the movement data analysis landscape, and we anticipate that its adoption will drive transformative ad- vancements and novel applications, making movement analysis a well-equipped field within the realm of computer science.

# Appendix A

# Summary of Techniques in AugmenTRAJ

This appendix contains a summary of augmentation candidate selection and point-modification techniques available in AugmenTRAJ for a quick reference.

| Method Name | User Controlled Parameters | Description |
|---|---|---|
| Random Selection | Proportion of trajectories to be selected | Randomly select the proportion of trajectories specified by the user as augmentation candidates. |
| Proportional Selection | Proportion of trajectories to be selected from each class | From each representative class, select the proportion of trajectories as specified by the user as augmentation candidates. |
| Length Based Selection | Proportion of trajectories to be selected | Selection of the user-specified proportion of trajectories that are shortest in length. |
| Representative Selection | Cutoff percentage for a trajectory to be considered a representative trajectory, tolerance value for a statistic to be considered close to representative statistic. | Select trajectories that most closely represent the entire dataset with the user specified closeness- cutoff. |

Table A.1: Summary of Selection Methods Available in AugmenTRAJ

| Method Name | User Controlled Parameters | Description |
|---|---|---|
| On-Circle Modification | - | Create the new point by selecting a circular region around the original point and randomly selecting a new point on the circumference of the circular region. |
| In-Circle Modification | - | Create the new point by selecting a circular region around the original point and randomly selecting a new point at a random distance within the circu- lar region and in a random direction. |
| Point Stretching Modification | Technique of stretching, max stretch (in meters) from the original point. | Create a new point at a user specified distance and in the user specified direction. |
| Point Dropping Modification | Probability of dropping a point | Drop points from the original probability with a user given probability. |

Table A.2: Summary of Point Modification Methods Available in AugmenTRA

# Appendix B

# Comparison of Base Strategy Metrics with Maximum Metrics Obtained with Augmentation

In this appendix, we will showcase the bar plots for result comparisons that we dis- cussed in Chapter 4. The box plots are a complete seed and model-wise breakdown of the data summarized in Table B.1. For each seed, it can be seen that the metrics are generally improving and in some cases up-to 51% as we discussed.

| Seed | Model (Classifier) | Base Accuracy | Base F1-Score | Maximum Accuracy | Maximum F1-Score | Maximum Accuracy Strategy | Maximum F1-Score Strategy |
|---|---|---|---|---|---|---|---|
| 781 | Gradient Boosting | 0.50 | 0.43 | 0.75 | 0.75 | balanced-in | balanced-in |
| 899 | Decision Tree | 0.50 | 0.50 | 0.88 | 0.88 | balanced-drop | balanced-drop |
| 1058 | Gradient Boosting | 0.62 | 0.48 | 1.00 | 1.00 | balanced-drop | balanced-drop |
| 1415 | Decision Tree | 0.75 | 0.77 | 1.00 | 1.00 | random-selected-on | random-selected-on |
| 1971 | Random Forest | 0.50 | 0.50 | 0.75 | 0.71 | balanced-in | balanced-in |
| 2097 | Random Forest | 0.38 | 0.38 | 0.62 | 0.63 | balanced-drop | balanced-drop |
| 2643 | Decision Tree | 0.62 | 0.56 | 1.00 | 1.00 | fewest-selected-on | fewest-selected-on |
| 2862 | Random Forest | 0.75 | 0.79 | 1.00 | 1.00 | balanced-drop | balanced-drop |
| 2884 | Decision Tree | 0.75 | 0.75 | 0.88 | 0.87 | balanced-drop | balanced-drop |
| 3589 | Decision Tree | 0.62 | 0.61 | 0.88 | 0.88 | balanced-in | balanced-in |
| 3832 | Decision Tree | 0.50 | 0.53 | 0.88 | 0.88 | fewest-selected-drop | fewest-selected-drop |
| 3846 | Decision Tree | 0.62 | 0.60 | 0.62 | 0.60 | balanced-drop | balanced-drop |
| 4944 | Decision Tree | 0.62 | 0.56 | 1.00 | 1.00 | fewest-selected-drop | fewest-selected-drop |
| 5923 | Gradient Boosting | 0.62 | 0.56 | 0.88 | 0.87 | balanced-on | balanced-on |
| 6406 | Gradient Boosting | 0.62 | 0.58 | 0.88 | 0.88 | balanced-drop | balanced-drop |
| 6939 | Gradient Boosting | 0.75 | 0.75 | 0.88 | 0.88 | balanced-stretch | balanced-stretch |
| 7932 | Gradient Boosting | 0.75 | 0.77 | 1.00 | 1.00 | balanced-stretch | balanced-stretch |
| 7950 | Random Forest | 0.38 | 0.20 | 0.75 | 0.75 | balanced-drop | balanced-drop |
| 9265 | Gradient Boosting | 0.50 | 0.57 | 0.75 | 0.79 | balanced-stretch | balanced-stretch |
| 9375 | Gradient Boosting | 0.50 | 0.57 | 1.00 | 1.00 | random-selected-on | random-selected-on |

Table B.1: Summary of Best Results Obtained For Each Seed Using AugmenTRAJ For the Geolife[43] Dataset

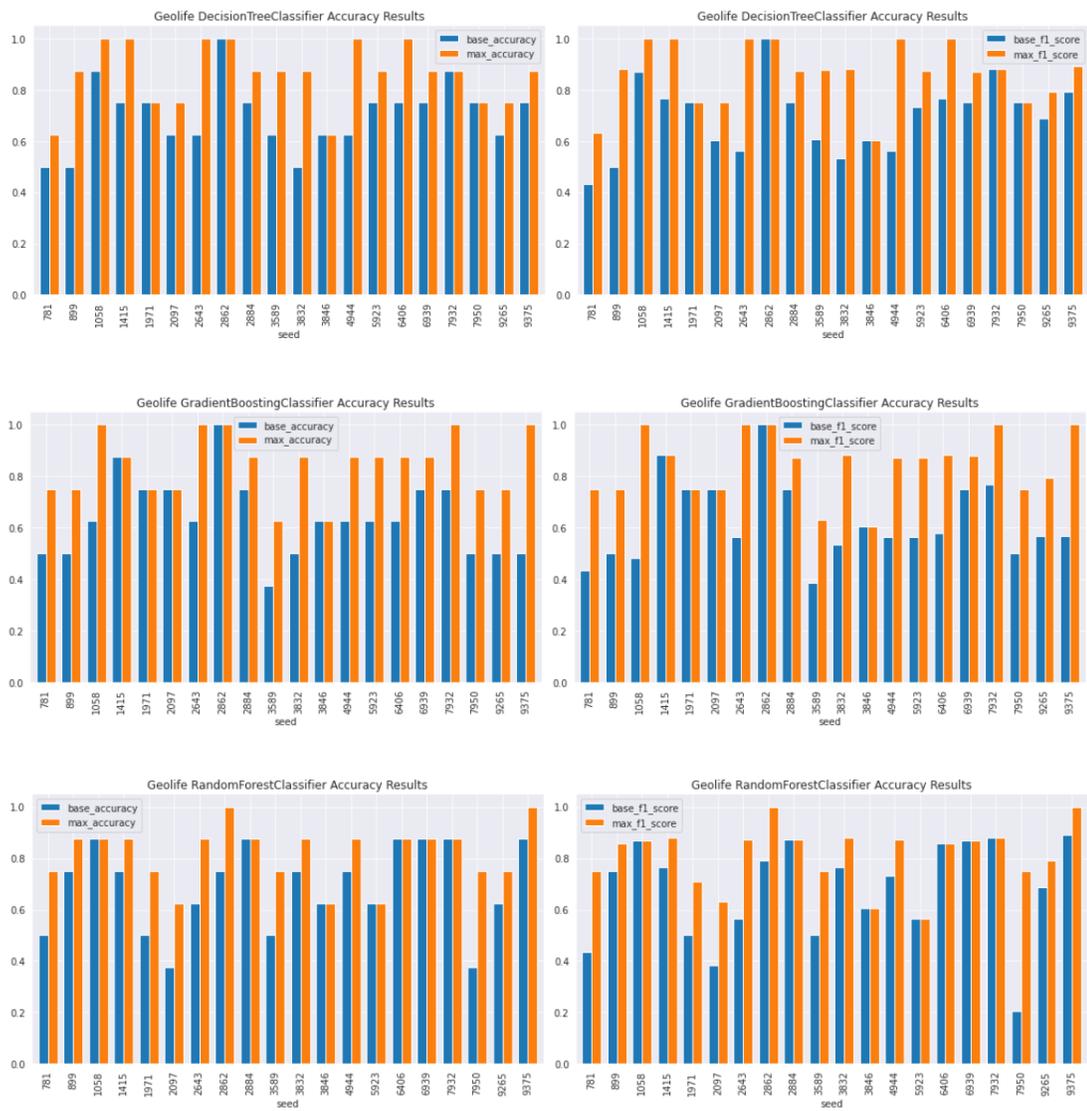

Figure B.1: Seed-wise Comparison between Base metrics and Maximum metrics for Geolife Dataset

| Seed | Model (Classifier) | Base Accuracy | Base F1-Score | Maximum Accuracy | Maximum F1-Score | Maximum Accuracy Strategy | Maximum F1-Score Strategy |
|---|---|---|---|---|---|---|---|
| 781 | SVC | 0.59 | 0.58 | 0.65 | 0.64 | fewest-selected-drop | fewest-elected-drop |
| 899 | SVC | 0.67 | 0.64 | 0.73 | 0.73 | balanced-drop | balanced-drop |
| 1058 | SVC | 0.67 | 0.62 | 0.75 | 0.75 | fewest-selected-stretch | fewest-selected-stretch |
| 1415 | SVC | 0.75 | 0.73 | 0.82 | 0.82 | fewest-selected-drop | fewest-selected-drop |
| 1971 | Gradient Boosting | 0.92 | 0.92 | 0.94 | 0.94 | representative-selected-in | representative-selected-in |
| 2097 | Decision Tree | 0.92 | 0.93 | 1.00 | 1.00 | proportional-selected-drop | proportional-selected-drop |
| 2643 | SVC | 0.73 | 0.71 | 0.76 | 0.76 | fewest-selected-in | fewest-selected-in |
| 2862 | SVC | 0.73 | 0.70 | 0.80 | 0.80 | fewest-selected-in | fewest-selected-in |
| 2884 | Decision Tree | 0.86 | 0.86 | 0.94 | 0.94 | random-selected-stretch | random-selected-stretch |
| 3589 | SVC | 0.69 | 0.66 | 0.75 | 0.74 | fewest-selected-drop | fewest-selected-drop |
| 3832 | Decision Tree | 0.86 | 0.86 | 0.92 | 0.92 | fewest-selected-on | proportional-selected-stretch |
| 3846 | Gradient Boosting | 0.90 | 0.89 | 0.92 | 0.92 | fewest-selected-drop | fewest-selected-drop |
| 4944 | Gradient Boosting | 0.88 | 0.88 | 0.94 | 0.94 | fewest-selected-on | fewest-selected-on |
| 5923 | SVC | 0.67 | 0.63 | 0.73 | 0.73 | balanced-in | balanced-in |
| 6406 | Decision Tree | 0.80 | 0.80 | 0.90 | 0.90 | representative-selected-stretch | representative-selected-stretch |
| 6939 | Decision Tree | 0.86 | 0.87 | 0.96 | 0.96 | fewest-selected-on | fewest-selected-on |
| 7932 | SVC | 0.63 | 0.61 | 0.69 | 0.70 | balanced-drop | fewest-selected-stretch |
| 7950 | Decision Tree | 0.84 | 0.84 | 0.88 | 0.88 | random-selected-in | random-selected-in |
| 9265 | SVC | 0.67 | 0.65 | 0.80 | 0.80 | balanced-drop | balanced-drop |
| 9375 | SVC | 0.65 | 0.62 | 0.73 | 0.73 | fewest-selected-drop | fewest-selected-drop |

Table B.2: Summary of Best Results Obtained For Each Seed Using AugmenTRAJ For the Starkey[39] Dataset

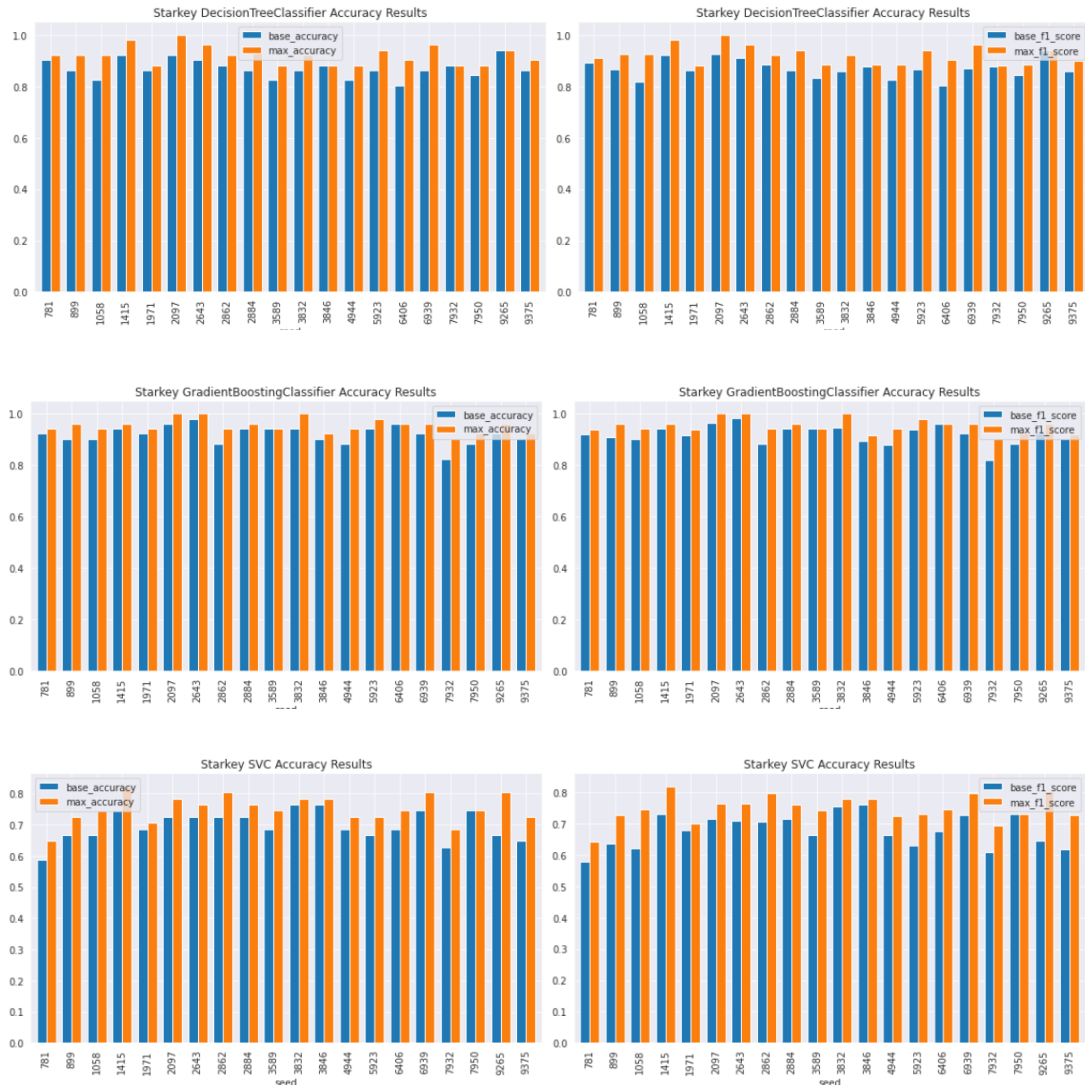

Figure B.2: Seed-wise Comparison between Base metrics and Maximum metrics for Starkey [39] Dataset

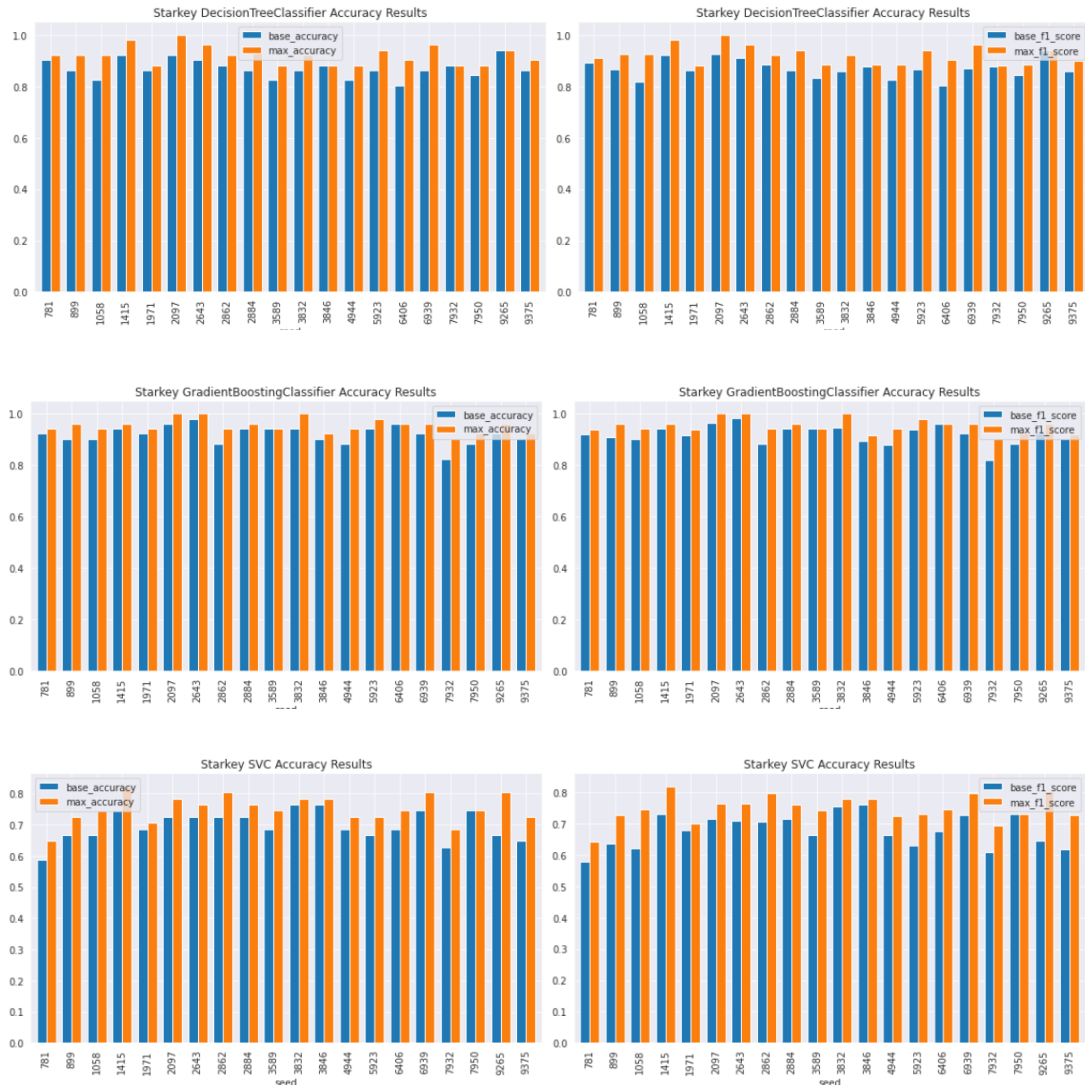

Figure B.2: Seed-wise Comparison between Base metrics and Maximum metrics for Starkey [39] Dataset

| Seed | Model (Classifier) | Base Accuracy | Base F1-Score | Maximum Accuracy | Maximum F1-Score | Maximum Accuracy Strategy | Maximum F1-Score Strategy |
|---|---|---|---|---|---|---|---|
| 781 | Decision Tree | 0.75 | 0.78 | 0.92 | 0.90 | balanced-drop | balanced-drop |
| 899 | Gradient Boosting | 0.88 | 0.82 | 0.92 | 0.90 | proportional-selected-drop | proportional-selected-drop |
| 1058 | Gradient Boosting | 0.88 | 0.90 | 0.96 | 0.94 | balanced-in | balanced-in |
| 1415 | Decision Tree | 0.88 | 0.89 | 0.96 | 0.96 | fewest-selected-drop | fewest-selected-drop |
| 1971 | Decision Tree | 0.76 | 0.76 | 0.92 | 0.90 | random-selected-in | random-selected-in |
| 2097 | Decision Tree | 0.79 | 0.77 | 0.92 | 0.90 | random-selected-stretch | random-selected-stretch |
| 2643 | Decision Tree | 0.92 | 0.90 | 0.96 | 0.96 | fewest-selected-on | fewest-selected-on |
| 2862 | Decision Tree | 0.75 | 0.80 | 0.92 | 0.92 | proportional-selected-in | proportional-selected-in |
| 2884 | SVC | 0.92 | 0.88 | 0.96 | 0.95 | balanced-on | balanced-on |
| 3589 | Decision Tree | 0.84 | 0.88 | 0.92 | 0.94 | fewest-selected-drop | proportional-selected-drop |
| 3832 | Decision Tree | 0.83 | 0.81 | 0.88 | 0.88 | proportional-selected-stretch | proportional-selected-stretch |
| 3846 | Decision Tree | 0.80 | 0.81 | 0.92 | 0.91 | fewest-selected-drop | fewest-selected-drop |
| 4944 | Gradient Boosting | 0.84 | 0.84 | 0.96 | 0.96 | representative-selected-on | representative-selected-on |
| 5923 | SVC | 0.92 | 0.88 | 0.96 | 0.95 | balanced-on | balanced-on |
| 6406 | Decision Tree | 0.80 | 0.77 | 0.84 | 0.83 | balanced-stretch | balanced-stretch |
| 6939 | Gradient Boosting | 0.88 | 0.89 | 0.96 | 0.95 | proportional-selected-drop | proportional-selected-drop |
| 7932 | Gradient Boosting | 0.76 | 0.73 | 0.88 | 0.85 | random-selected-drop | random-selected-drop |
| 7950 | Decision Tree | 0.76 | 0.73 | 0.84 | 0.84 | representative-selected-stretch | representative-selected-stretch |
| 9265 | Gradient Boosting | 0.92 | 0.92 | 1.00 | 1.00 | balanced-in | balanced-in |
| 9375 | Decision Tree | 0.84 | 0.86 | 0.96 | 0.95 | proportional-selected-drop | proportional-selected-drop |

Table B.3: Summary of Best Results Obtained For Each Seed Using AugmenTRAJ For the Traffic[14] Dataset

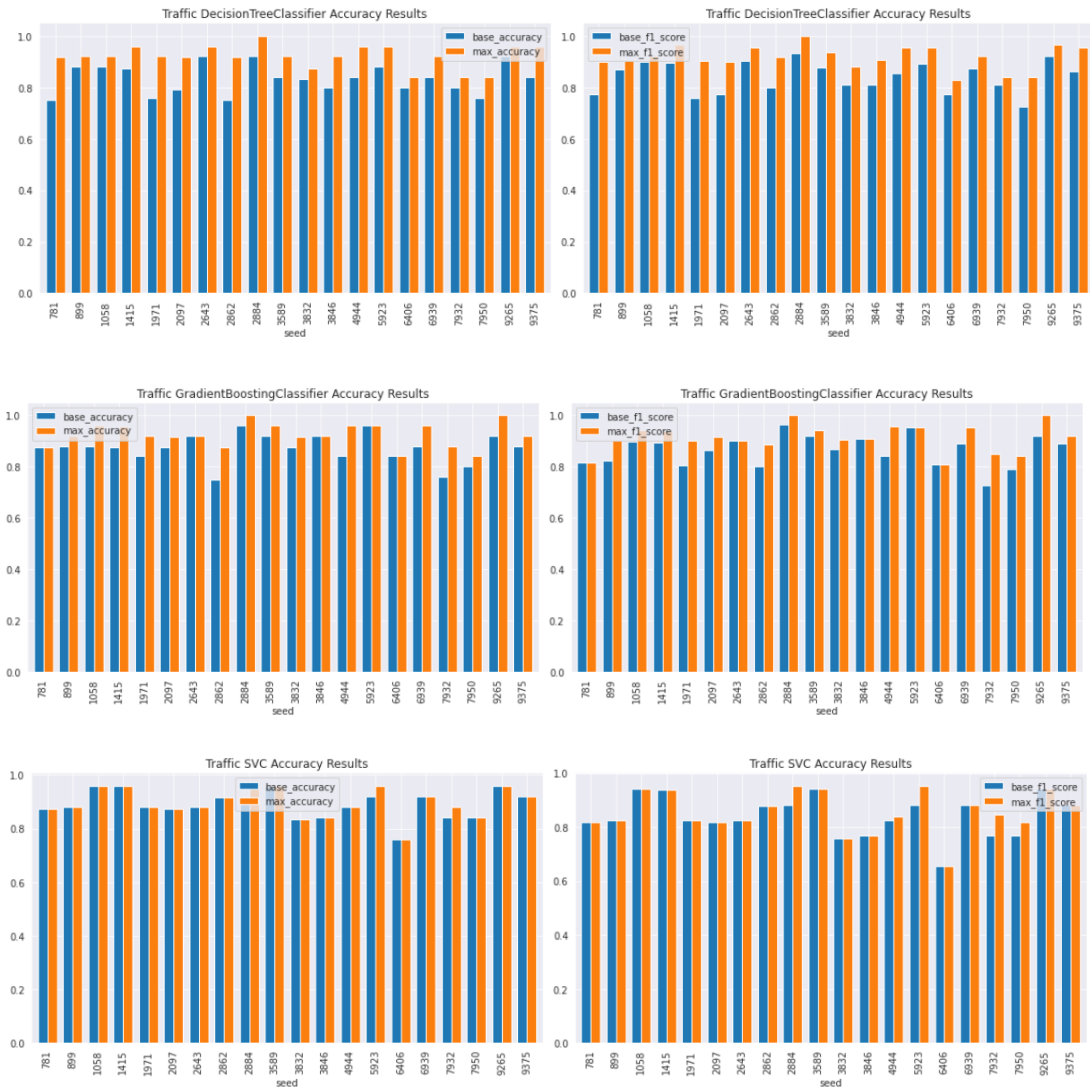

Figure B.3: Seed-wise Comparison between Base metrics and Maximum metrics for Traffic [14] Datase

[42] L. Zhao and G. Shi. A trajectory clustering method based on douglas-peucker compression and density for marine traffic pattern recognition. *Ocean Engineer- ing*, 172:456–467, 2019.

[43] Y. Zheng, H. Fu, X. Xie, W.-Y. Ma, and Q. Li. *Geolife GPS trajectory dataset - User Guide*, geolife gps trajectories 1.1 edition, July 2011.